\documentclass[twocolumn,pdflatex,sn-nature]{sn-jnl}% Style for submissions to Nature 
 
\usepackage{geometry}
\usepackage{graphicx}%
\usepackage{multirow}%
\usepackage{amsmath,amssymb,amsfonts}%
\usepackage{amsthm}%
\usepackage{mathrsfs}%
\usepackage[title]{appendix}%
\usepackage{xcolor}%
\usepackage{colortbl}
\usepackage{textcomp}%
\usepackage{manyfoot}%
\usepackage{booktabs}%
\usepackage{algorithm}%
\usepackage{algorithmicx}%
\usepackage{algpseudocode}%
\usepackage{listings}%
\usepackage[UTF8]{ctex}
\renewcommand{\figurename}{Figure}
\renewcommand{\abstractname}{Abstract}
\theoremstyle{thmstyleone}%
\usepackage{amsmath}
\usepackage{amssymb}
\usepackage{cuted}

\theoremstyle{thmstyletwo}%

\theoremstyle{thmstylethree}%
\makeatletter
\newcounter{subsubsubsection}[subsubsection]

\newcommand{\subsubsubsection}{\@startsection{subsubsubsection}{4}{\z@}%
  {-3.25ex\@plus -1ex \@minus -.2ex}{1.5ex \@plus .2ex}{\normalfont\normalsize\bfseries}}
\makeatother
\newenvironment{mycustomequation}
    {\fontsize{10.0pt}{10pt}\selectfont\begin{equation}}
    {\end{equation}}

\begin{document}

\title[Article Title]{Neural Dynamics-Informed Pre-trained Framework for Personalized Brain Functional Network Construction}

\author[1]{\fnm{Hongjie} \sur{Jiang}}
\author[1]{\fnm{Yifei} \sur{Tang}}
\author*[1]{\fnm{Shuqiang} \sur{Wang}}\email{sq.wang@siat.ac.cn}

\affil[1]{\orgdiv{Shenzhen Institutes of Advanced Technology}, \orgname{Chinese Academy of Sciences}, \orgaddress{\city{Shenzhen}, \country{China}}}

\abstract{Brain activity is intrinsically a neural dynamic process constrained by anatomical space. This leads to significant variations in spatial distribution patterns and correlation patterns of neural activity across variable and heterogeneous scenarios. However, dominant brain functional network construction methods, which relies on pre-defined brain atlases and linear assumptions, fails to precisely capture varying neural activity patterns in heterogeneous scenarios. This limits the consistency and generalizability of the brain functional networks constructed by dominant methods. Here, a neural dynamics-informed pre-trained framework is proposed for personalized brain functional network construction. The proposed framework extracts personalized representations of neural activity patterns in heterogeneous scenarios. Personalized brain functional networks are obtained by utilizing these representations to guide brain parcellation and neural activity correlation estimation. Systematic evaluations were employed on 18 datasets across tasks, such as virtual neural modulation and abnormal neural circuit identification. Experimental results demonstrate that the proposed framework attains superior performance in heterogeneous scenarios. Overall, the proposed framework challenges the dominant brain functional network construction method.}

\maketitle

\section*{Introduction}\label{sec1}

Brain functional networks reflect patterns of information flow between brain regions and have become an indispensable analytical tool for understanding complex neural activity \cite{fc4infocomm}. Brain functional networks are widely applied in fields such as revealing brain mechanisms\cite{fc4brainmechanism3}, optimizing neural modulation strategies\cite{fc4modulation1}, and analyzing abnormal neural circuits \cite{fc4diseasemechanism2}. Thus, an efficient brain functional network construction framework is essential for supporting these application scenarios.

Recently, Cruces et al. proposed Micapipe framework. This framework integrates anatomical brain atlases, such as Desikan-Killiany, alongside functional brain atlases like Schaefer. It constructs brain functional networks based on Pearson correlation coefficients. This framework has been proved effective on high-resolution MICs and MSC fMRI datasets\cite{Micapipe}. Additionally, Siemon et al. developed the CATO framework. This framework integrates anatomical brain atlases such as Von Economo and Koskinas. It supports three methods for estimating neural activity correlations: Pearson correlation, partial correlation, and covariance. This framework has been proved applicable to the HCP dataset of healthy adults \cite{CATO}. However, functional units are task-\cite{functional_unit_is_task_state_subject1}, state-\cite{age_groups}, and subject-dependent\cite{functional_unit_is_task_state_subject1, functional_unit_is_task_state_subject2}. The above dominant method in brain functional network construction based on the "pre-defined atlas and linear assumption"\cite{CATO, Micapipe, gretna, XCP-D} fails to effectively adapt to the spatial distribution variation of neural activity in variable and heterogeneous scenarios. This limits the consistency and generalizability performance of brain functional networks constructed by dominant methods. Specifically, characterized as a neural dynamic process inherently constrained by anatomical space~\cite{anatomic_constrain}, brain activity in variable and heterogeneous real-world scenarios is influenced by factors such as subjects' age group~\cite{age_groups}, brain disorder type~\cite{ASD_Scenario, AD_Scenario, ADHD_Scenario, PD_Scenario, MDD_Scenario, Different_Disease_Different_Atlas}, race (e.g., linguistic differences)~\cite{languages}, and image acquisition strategies~\cite{scan_duration1, scan_duration2}. Consequently, the observed neural activity patterns exhibit significant variations(variations in both spatial distribution patterns and correlation patterns). Faced with such variations in neural activity patterns, the dominant brain functional network construction method, which relies on pre-defined brain atlases derived from ex vivo specimens and group averages as well as linear estimations of neural activity correlations, inevitably overlooks the differences in neural activity spatial distribution patterns. For example, Cui et al. demonstrates that neural activity spatial distribution patterns in the association cortex differ significantly across age groups\cite{cui2020individual}. The pre-defined brain atlases widely used in current methods, such as Desikan-Killiany and Schaefer\cite{schaefer2018local, dk_atlas}, cannot be adaptively adjusted and therefore fail to capture the differences. Furthermore, linear estimations of neural activity correlations\cite{CATO, Micapipe} fail to effectively characterize the varying correlation patterns in such scenarios. Consequently, a personalized brain functional network construction framework is critical to address the challenge of precisely capturing neural activity patterns in heterogeneous scenarios.

Here, a neural dynamics-informed pre-trained framework is proposed for personalized brain functional network construction. The proposed framework aims to capture neural activity patterns in heterogeneous scenarios by using personalized representations of neural activity patterns to guide brain parcellation and correlation estimation, thereby constructing personalized brain functional networks. First, a foundation model is pre-trained on large-scale functional magnetic resonance imaging (fMRI) data to provide the basis for capturing neural activity pattern representations in heterogeneous scenarios. Second, to enable the proposed framework with the ability to extract personalized neural activity pattern representations in specific scenarios, the foundation model is fine-tuned with neural dynamics information. Third, the personalized brain parcellation is obtained by utilizing these representations to guide the adaptive brain parcellation module. Meanwhile, driven by neural dynamics, these representations provide dynamic information that cannot be obtained from fMRI data alone. This ensures that the brain parcellation maintains anatomical plausibility while more effectively capturing changes in neural activity spatial distribution patterns. Finally, based on this brain parcellation, the neural activity correlation estimation module is guided by personalized representations. This guidance enables the personalized brain functional networks construction by estimating neural activity correlation patterns in heterogeneous scenarios.

Personalized brain functional networks were constructed using 18 datasets and evaluated across heterogeneous scenarios from multiple perspectives. These 18 datasets include fMRI data across three age groups (children, adolescents, and the elderly)\cite{age_groups}, five brain disorders (Alzheimer’s Disease (AD), Attention Deficit Hyperactivity Disorder (ADHD), Parkinson’s Disease (PD), Major Depressive Disorder (MDD), and Autism Spectrum Disorder (ASD))\cite{AD_Diagnosis, ADHD_Diagnosis, ASD_Diagnosis, PD_Diagnosis, MDD_Diagnosis}, and various acquisition strategies (long scan duration, short scan duration). Experimental results demonstrate that the personalized brain functional networks constructed by the proposed framework are more adaptable to heterogeneous scenarios than those constructed by the baseline method. Specifically, compared with the brain functional networks constructed by the baseline method, personalized brain functional networks (1) maintain enhanced consistency\cite{PDiv}; (2) boost classification accuracy when applied to multiple tasks such as brain disorder diagnosis, motor imagery decoding\cite{fc_motor_imagery_decode}, and individual identification\cite{individual_indenti}; (3) enable the identification of more effective neural modulation targets\cite{modulation_target_identification_by_fc}, and (4) facilitate the identification of highly consistent abnormal neural circuits across datasets with varying acquisition strategies. Meanwhile, an open-source Python tool for personalized brain functional network construction has been released. This study establishes a paradigm for constructing personalized brain functional networks.

\section*{Results}\label{sec2}

\begin{figure*}[htbp]
    \centering
    \includegraphics[width=\textwidth]{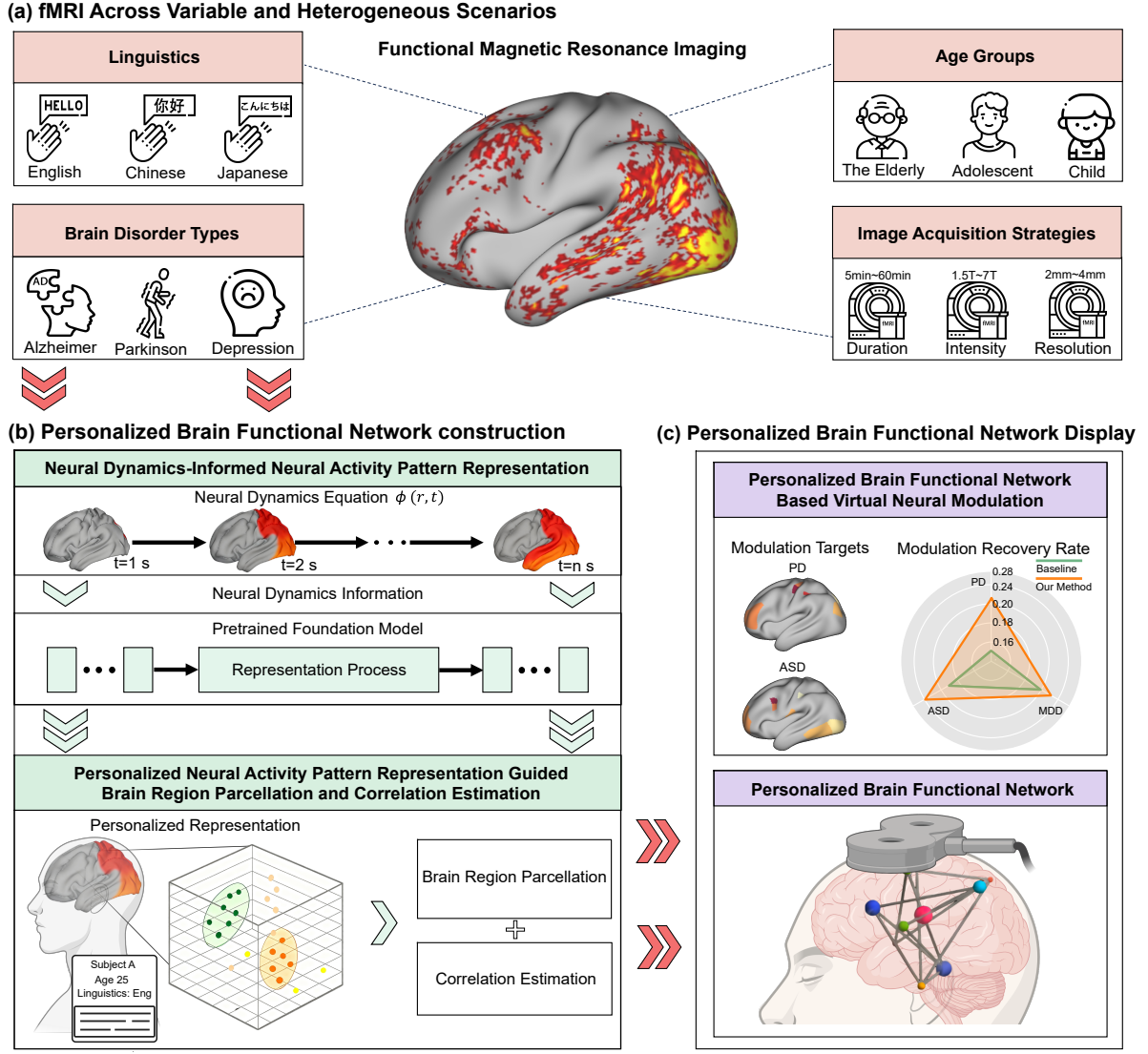}
    \caption{\textbf{Overview of the Personalized Brain Functional Network Construction Framework.} \textbf{a}, Personalized brain functional network construction framework can be adapted to fMRI acquired across variable and heterogeneous scenarios such as subjects' age groups, brain disorder types, linguistics, and image acquisition strategies. \textbf{b}, The proposed framework begins with a foundation model pretrained on large-scale fMRI data. The model is then fine-tuned with neural dynamics information to extract personalized representations from neural activities. These representations are used to guide the brain region parcellation and the correlation estimation to construct the personalized brain functional network.} %
    \label{fig:1}
\end{figure*}

\begin{figure*}[htbp]
    \centering
    \includegraphics[width=\textwidth]{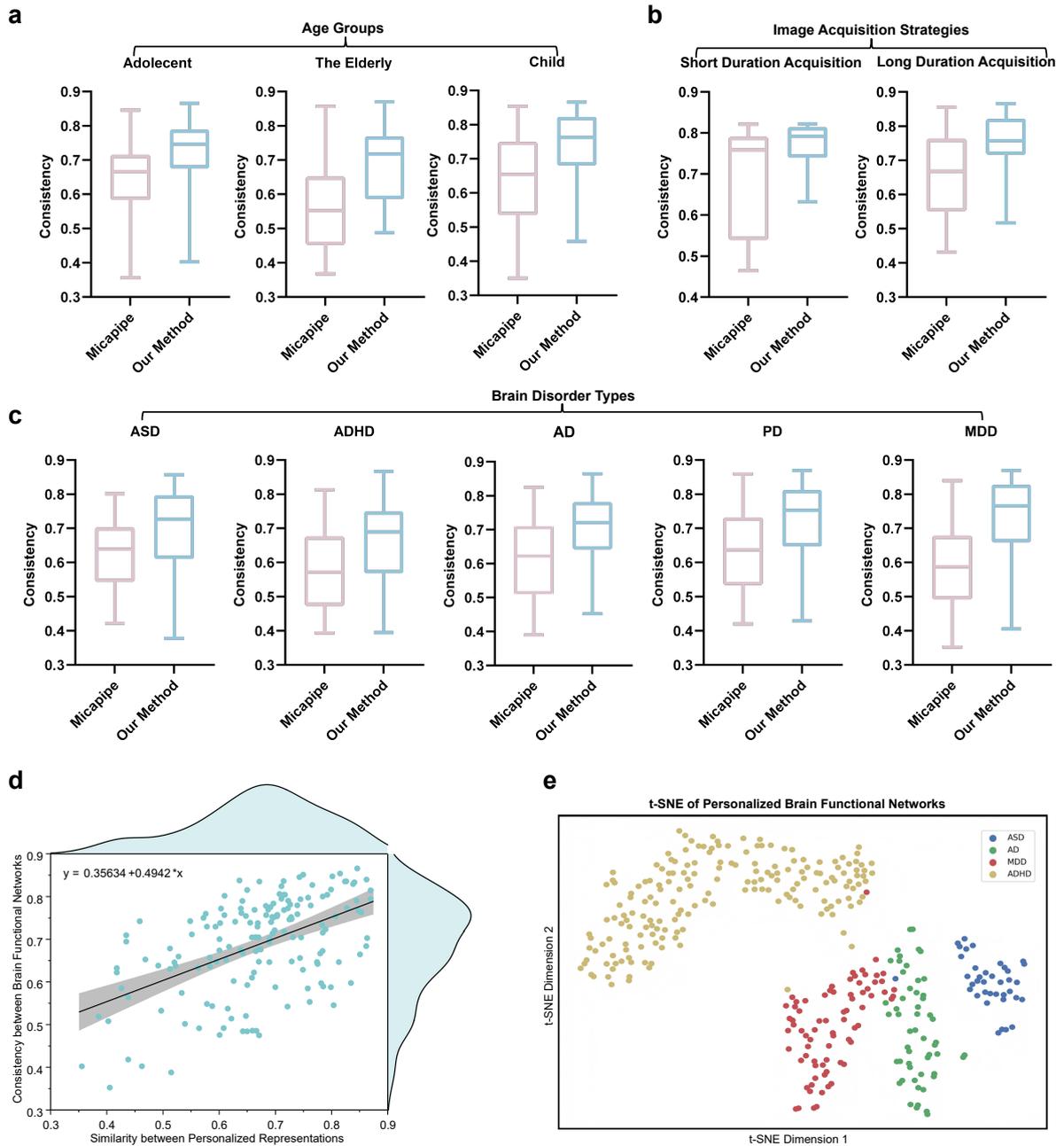}
    \caption{\textbf{Consistency evaluation of personalized brain functional networks.} \textbf{a–c}, Box plots of PDiv distributions comparing the personalized brain functional network constructed by our method (blue) with brain functional network constructed by the baseline Micapipe method (pink) across heterogeneous scenarios. Results demonstrate that our method achieves significantly higher median PDiv values across (a) varying age groups , (b) image acquisition strategies , and (c) brain disorder types, consistently outperforming the baseline method. \textbf{d}, Regression analysis of inter-representation similarity versus brain functional network consistency. Results demonstrate that personalized representations are the underlying mechanism for enhanced brain functional network consistency. \textbf{e}, t-SNE visualization of personalized brain functional network features for different disorders (ASD, AD, MDD, ADHD). Results demonstrate that the high consistency of the personalized brain functional network is achieved with ignoring the distinct discriminative features of neural activity patterns.}
    \label{fig:2}
\end{figure*}

\begin{figure*}[htbp]
    \centering
    \includegraphics[width=1.0\textwidth]{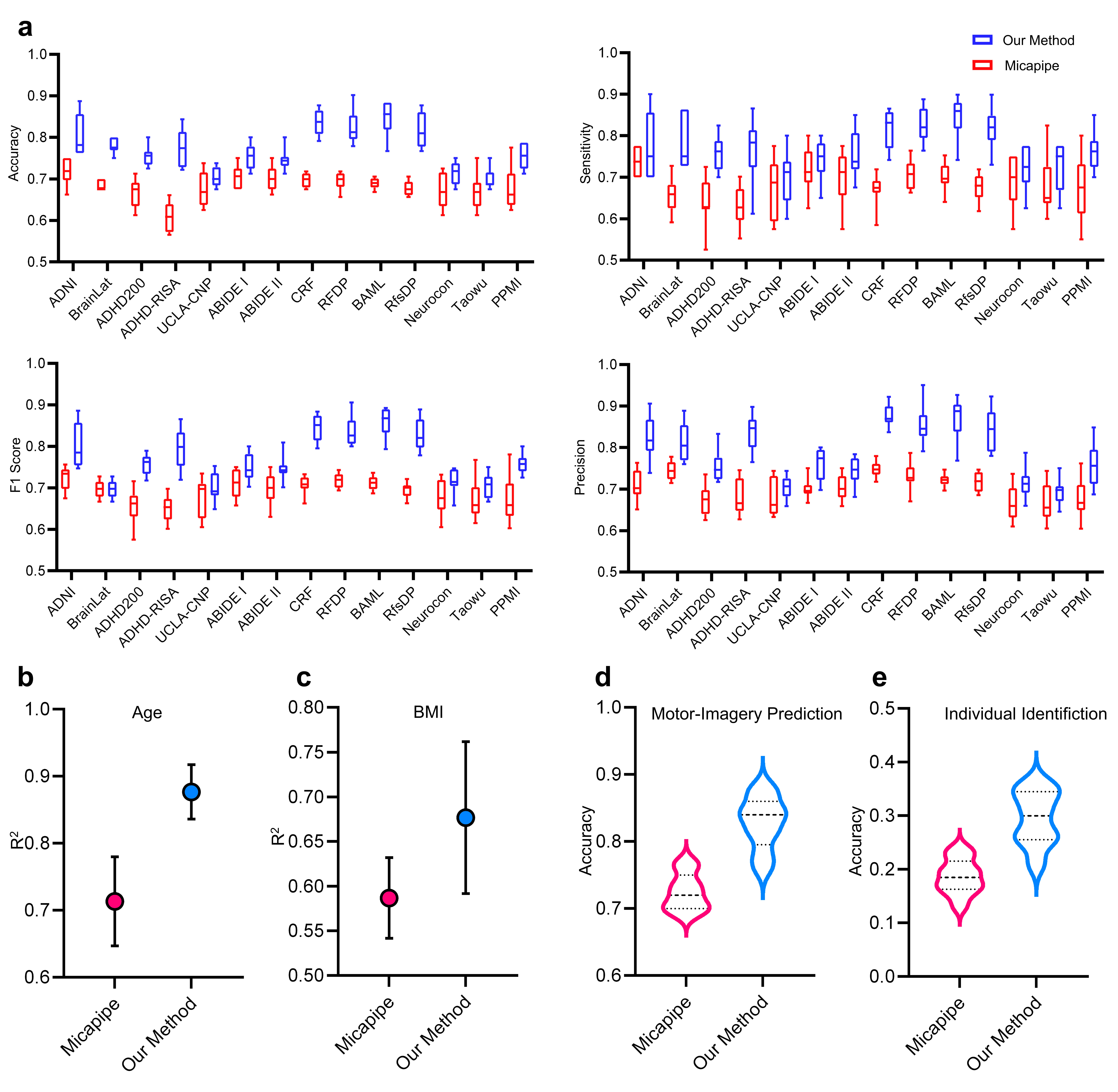}
    \caption{\textbf{Evaluation of personalized brain functional networks in multiple diagnosis and prediction tasks.} \textbf{a}, Box plots comparing diagnostic performance metrics (Accuracy, Sensitivity, Precision, F1) across multiple brain disorder (AD, PD, MDD, ADHD, ASD). Results demonstrate that our method consistently outperforms the baseline method across all diagnosis tasks. \textbf{b, c}, Comparison of $R^2$ for physiological index prediction. Results demonstrate that our method achieves significantly higher predictive accuracy for (b) age  and (c) BMI. \textbf{d, e}, Performance evaluation for (d) motor imagery decoding and (e) individual identification. Results demonstrate that our method significantly surpasses the baseline method.}
    \label{fig:3}
\end{figure*}

\begin{figure*}[htbp]
    \centering
    \includegraphics[width=\textwidth]{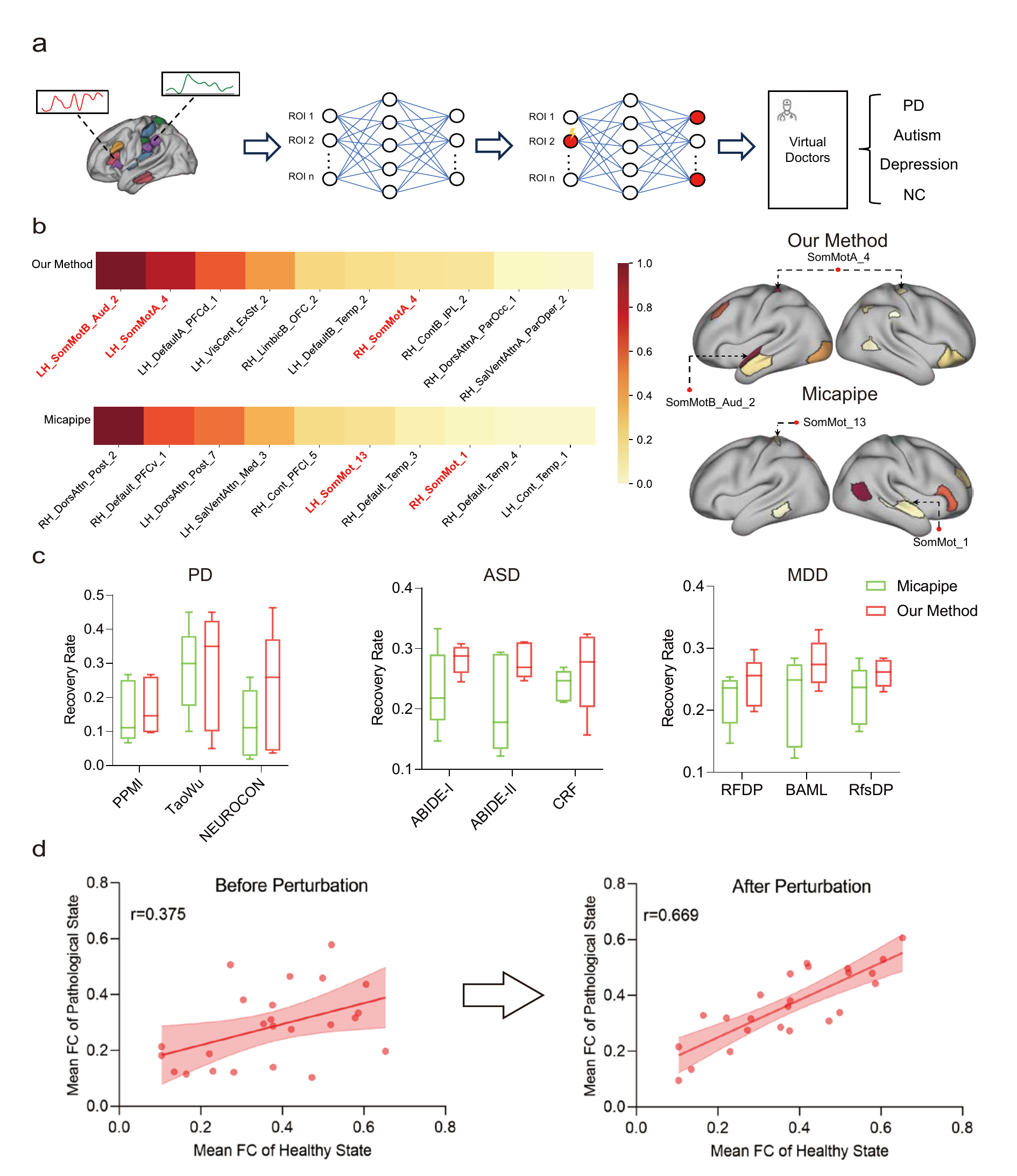}
    \caption{\textbf{Evaluation of personalized brain functional networks via virtual neural modulation across three brain disorders.} \textbf{a}, Schematic illustration of the virtual neural modulation. The process begins by simulating brain activity states using a neural network. Next, perturbations are applied to the optimal targets identified by SHAP analysis. Finally, a "Virtual Doctors" classifier is applied to diagnose whether patients have recovered. \textbf{b}, Visualization of neural modulation targets for PD based on the TaoWu dataset. The heatmap on the left shows the importance rankings of different brain regions as neural modulation targets. The key high-priority targets are highlighted in red. The brain surface maps on the right illustrate the spatial distribution differences between the core targets identified by our method and the baseline method. Results demonstrate that our method identifies targets more consistent with empirical clinical evidence compared with the baseline method. \textbf{c}, The box plots show recovery rates after virtual perturbation across nine independent datasets for PD, ASD, and MDD. Results demonstrate that the proposed method achieves significantly higher recovery rates across diverse datasets compared with the baseline method. \textbf{d}, The scatter plot shows the change in correlation of mean functional connectivity (Mean FC) between the MDD patient group and healthy controls. The left panel depicts the state before perturbation, and the right panel shows the state after perturbation. Results demonstrate that virtual modulation effectively transform pathological brain activity towards healthy patterns.}
    \label{fig:4}
\end{figure*}

\begin{figure*}[htbp]
    \centering
    \includegraphics[width=\textwidth]{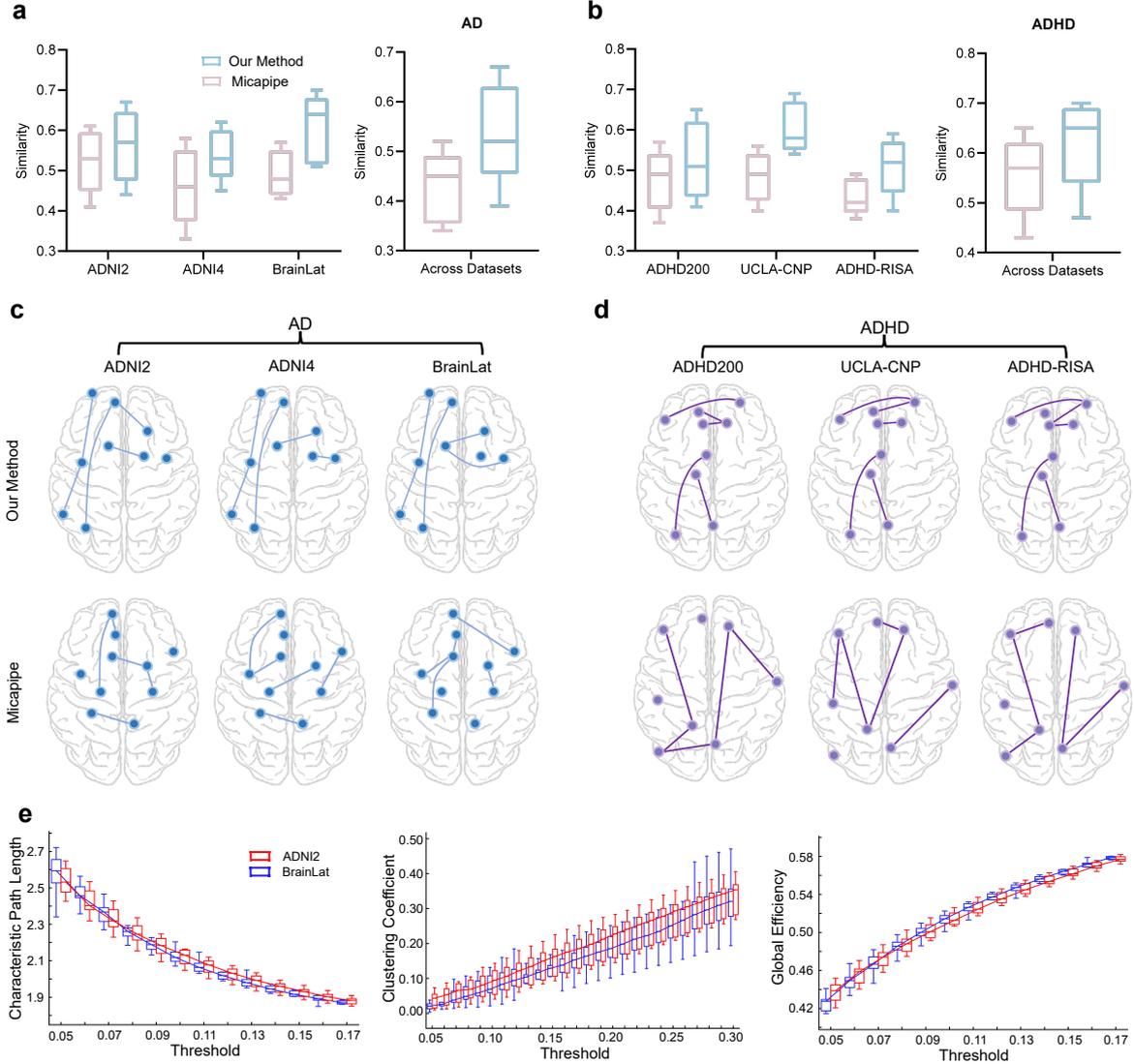}
    \caption{\textbf{Abnormal neural circuit identification and consistency analysis across datasets with different image acquisition strategies.} \textbf{a}, Box plots illustrating cosine similarity between abnormal neural circuits identified in AD datasets with different acquisition strategies. Results demonstrate that the abnormal neural circuits identified by our method exhibit significantly higher consistency across AD datasets with different acquisition strategies compared to the baseline method. \textbf{b}, Box plots illustrating cosine similarity between abnormal neural circuits identified in ADHD datasets with different acquisition strategies. Results demonstrate that the abnormal neural circuits identified by our method exhibit significantly higher consistency across ADHD datasets with different acquisition strategies compared to the baseline method. \textbf{c, d}, Topographical visualization of the top 5 highest-weighted abnormal connections for (c) AD and (d) ADHD datasets with different acquisition strategies. Results demonstrate the robust visual consistency of neural circuits identified by our method across datasets with different acquisition strategies. \textbf{e}, Comparative analysis of neural circuit complex network metrics as a function of connectivity threshold, focusing on characteristic path length, clustering coefficient, and global efficiency. Results demonstrate highly consistent trends between datasets with different acquisition strategies.}
    \label{fig:5}
\end{figure*}

\subsection*{Consistency evaluation of the personalized brain functional network}\label{subsec1}
To evaluate the performance of the personalized brain functional network construction framework under variable and heterogeneous scenarios, Portrait Divergence (PDiv) was adopted as the metric\cite{PDiv}. This metric assesses whether the personalized brain functional network maintains higher consistency for the same subject in a short, intervention-free session compared to the brain functional network constructed by the baseline method (Micapipe)\cite{Micapipe}. First, the statistical distributions of PDiv values were systematically examined across three dimensions: age groups (children, adolescent, and the elderly), acquisition strategies (long scan duration, short scan duration), and brain disorder types (AD, PD, MDD, etc.). Second, regression analysis was employed to further investigate the relationship between personalized representations and the consistency of personalized brain functional networks. Finally, to ensure that the high consistency of the personalized brain functional networks is not obtained at the cost of ignoring the ability to reflect critical differentiation information across neural activity patterns, validation was employed to determine whether key discriminative features of neural activity patterns are successfully reflected by the personalized brain functional network.

To evaluate the consistency of personalized brain functional networks under heterogeneous scenarios, PDiv distributions from both our method-constructed and baseline method-constructed brain functional networks were systematically compared. As shown in Figs. 2a, 2b, and 2c, experimental results demonstrate that personalized brain functional networks significantly outperformed baseline method in consistency. First, across different age groups (Fig. \ref{fig:2}a), the median one minus PDiv values of personalized brain functional networks (0.7-0.8), significantly higher the baseline method (0.5-0.7). Second, under both varying acquisition strategies (Fig. \ref{fig:2}b) and different brain disorders (Fig. \ref{fig:2}c), the median one minus PDiv values of personalized brain functional networks (0.7-0.8) were significantly higher than the baseline method (0.55-0.75). To further investigate the underlying factors contributing to the high consistency of personalized brain functional networks, the correlation between consistency and inter-representation similarity was analyzed. As shown in Fig. \ref{fig:2}d, experimental results demonstrate that inter-representation similarity is significantly and positively correlated with the consistency of personalized brain functional networks. Specifically, the linear regression coefficient between inter-representation similarity and the consistency of personalized brain functional networks is 0.49. Additionally, the 95\% confidence interval of the regression line remains narrow. To verify that the high consistency of personalized brain functional networks is not obtained at the cost of losing critical differentiation information across neural activity patterns, t-SNE analysis was performed on personalized brain functional networks corresponding to different brain disorders. As shown in Fig. \ref{fig:2}e, experimental results demonstrate that high-consistency personalized brain functional networks retain differentiation of neural activity pattern under heterogeneous scenarios. t-SNE analysis for ASD, AD, ADHD, and MDD shown separated clusters for each disorder. Samples within each group clustered tightly. Overall, compared to the baseline method, personalized brain functional networks shown significantly higher consistency across heterogeneous scenarios. Quantitative analysis reveals that personalized representations are the inner mechanism driving the enhanced consistency. More importantly, high-consistency personalized brain functional networks do not lose differentiation information across neural activity patterns under heterogeneous scenarios.

\subsection*{Evaluation of personalized brain functional networks in diagnosis and prediction tasks.}\label{subsec2}
Systematic experiments were conducted to comprehensively evaluate the effectiveness of the personalized brain functional networks constructed by the proposed method in diagnosis and prediction tasks. Evaluations were conducted on tasks including brain disorder diagnosis (AD, PD, MDD, ADHD, ASD), individual identification, prediction of physiological indices (age/BMI)\cite{bio_index}, and motor imagery decoding.

First, the performance of the personalized brain functional network was validated in brain disorder diagnosis tasks. Performance of the diagnostic method based on personalized brain functional networks was compared against the baseline diagnostic method (the method based on brain functional networks constructed by the current method). As shown in Fig. \ref{fig:3}a, experimental results demonstrate that in brain disorder diagnosis tasks, the proposed method consistently outperformed the baseline diagnostic method across all datasets. Specifically, the the proposed method maintained a median accuracy of 0.73 to 0.90, significantly outperforming the baseline method that generally fell below 0.73. Across sensitivity, precision, and F1 score, the proposed method also maintained higher median values. Second, to validate performance in prediction tasks, physiological index prediction experiments were conducted. As shown in Figs. 3b and 3c, experimental results demonstrate that the proposed method more accurately captures metabolic-related neural activity features, achieving significantly higher Goodness of Fit (R2) than baselines. Specifically, for age prediction (Fig. \ref{fig:3}b), the method achieved a mean $R^2$ near 0.90 whereas the baseline was only 0.72. Similarly, in BMI prediction (Fig. \ref{fig:3}c), the proposed method achieved excellent R2 performance. Finally, in individual identification and motor-imagery decoding tasks (Figs. \ref{fig:3}d, \ref{fig:3}e), experimental results demonstrate that the proposed method significantly outperformed the baseline method. For motor imagery decoding (Fig. \ref{fig:3}d), the proposed method's median accuracy exceeded 0.80, compared to below 0.75 for the baseline. In individual identification (Fig. \ref{fig:3}e), the method significantly surpassed the baseline in accuracy. Overall, experimental results demonstrate that the proposed method obtained comprehensive superiority over the above tasks.

\subsection*{Evaluation of personalized brain functional networks via virtual neural modulation across three brain disorders. }\label{subsec3}
To evaluate the potential of personalized brain functional networks in neural modulation, virtual modulation experiments were conducted across nine datasets: PPMI, TaoWu, NEUROCON, ABIDE-I, ABIDE-II, CRF, RFDP, BAML, and RfsDP. The workflow of the virtual modulation experiment is illustrated in Fig. \ref{fig:4}a. First, neural activity was modeled to simulate brain dynamics. Second, perturbations were applied to the optimal targets identified by SHAP analysis\cite{modulation_target_identification_by_fc, SHAP}. Third, a virtual clinician assessed wheather the patient is recovered based on the post-perturbation brain activity. Finally, regression analysis was employed to evaluate the changes in the patient's brain activity patterns induced by the virtual modulation.

The performance of our method in localizing neural modulation targets was validated. As shown in Fig. \ref{fig:4}b, experimental results demonstrate that the method based on personalized brain functional networks identifies targets consistent with empirical modulation evidence more precisely than baseline method. Multiple previous studies have confirmed that the Somatomotor Network (SMN) is a key modulation target for alleviating PD symptoms. Studies by Elahi et al., Benninger et al., and Brittain et al. have respectively demonstrated that repetitive transcranial magnetic stimulation (rTMS), transcranial direct current stimulation (tDCS), and transcranial alternating current stimulation (tACS) targeting the SMN effectively ameliorate motor impairments or tremors in PD patients\cite{TMS_target,tDCS_target,tACS_target}. In TaoWu dataset, highly consistent with these findings, SMN exhibits a significant and reliable contribution in the proposed method. In the baseline method, the contributions of the targets in Dorsal Attention Network and Default Mode Network are more prominent, whereas the contribution of the SMN is relatively low. Existing works lack direct evidence supporting these networks as effective targets for alleviating PD symptoms. To further validate the efficacy of the targets, virtual modulation was conducted on the identified target regions and recovery rates were recorded. Furthermore, as illustrated in Extended Data Fig. 1 and Extended Data Fig. 2, our method also identified targets more aligned with empirical modulation experience than the baseline method across the PPMI, Neurocon, ABIDE-I, and ABIDE-II datasets. As illustrated in Fig. \ref{fig:4}c, experimental results demonstrate that virtual neural modulation based on the proposed method significantly outperformed the baseline method on recovery rates. Specifically, in the PD, ASD and MDD datasets, the median recovery rates obtained by proposed method improved by 8\%, 6.3\%, and 2\%, respectively, compared to baseline method. This provides robust evidence that targets identified via personalized brain functional networks remain effective across these diverse scenarios. To evaluate the efficacy of virtual modulation in modifying patient brain activity patterns, regression analysis was performed on the personalized brain functional networks before and after modulation. As illustrated in Fig. \ref{fig:4}d, experimental results demonstrate that virtual neural modulation based on the proposed method effectively normalizes brain activity in patients with MDD. Compared to the pre-modulation, the correlation coefficient between the post-modulation brain functional networks and the healthy-state networks increased by 0.29. Collectively, these experimental results demonstrate the substantial application potential of personalized brain functional networks for neural modulation.

\subsection*{Abnormal neural circuit identification and consistency analysis across datasets with different image acquisition strategies.}\label{subsec5}
To assess the consistency of abnormal neural circuits \cite{abnormal_neural_circuits_identification} identified by personalized brain functional networks across datasets with different acquisition strategies, experiments were conducted on six datasets with varying acquisition strategies: ADNI2, ADNI4, BrainLat, ADHD200, UCLA-CNP, and ADHD-RISA. The disease-specific abnormal neural circuit was obtained by calculating the difference between the average neural circuits of the healthy and disorder groups. The consistency of abnormal circuit identification results across datasets with different acquisition strategies was evaluated from three perspectives: similarity metrics, visual analysis, and complex network metrics.

First, cosine similarity was utilized as a quantitative metric to assess the consistency of abnormal circuit identification results across datasets with different acquisition strategies. As shown in Figs. 5a and 5b, experimental results demonstrate that compared to brain functional networks constructed by baseline methods, the abnormal neural circuit identification results based on personalized brain functional networks exhibit significantly higher cosine similarity across datasets with different acquisition strategies. Similarity assessments were conducted on three AD datasets: ADNI2 (resolution=3.8mm), ADNI4 (resolution=2.5mm), and BrainLat (resolution=3mm). Compared to baseline methods, the median cosine similarity of abnormal circuits identified based on personalized brain functional networks increased by 0.04, 0.07, and 0.16 on the ADNI2, ADNI4, and BrainLat datasets, respectively. Furthermore, the cosine similarity between abnormal circuits identified based on personalized brain functional networks across these three datasets improved by 0.07 compared to baseline methods. To validate the performance of personalized brain functional networks across different brain disorder types, experiments were further conducted on ADHD. Experimental results indicate that in the three ADHD datasets—ADHD200 (TR=2.25s), UCLA-CNP (TR=2s), and ADHD-RISA (TR=3s)—the cosine similarity between abnormal circuits identified by the proposed method was significantly improved compared to baseline methods. Second, to visually assess consistency across datasets with different acquisition strategies, the top 5 highest-weighted abnormal connections were visualized. As shown in Fig. \ref{fig:4}c and \ref{fig:4}d, experimental results demonstrate that compared to the baseline method, abnormal circuits identified by personalized brain functional networks exhibit higher consistency across AD datasets with different acquisition strategies. Similarly, in ADHD datasets with different acquisition strategies, the identified circuits demonstrated higher consistency than baseline methods. Finally, consistency across datasets with different acquisition strategies was further evaluated via complex network metrics. As shown in Fig. \ref{fig:5}e, experimental results demonstrate that abnormal circuits identified by personalized brain functional networks are highly consistent in these metrics across datasets. Characteristic path length, clustering coefficient, and global efficiency \cite{complex_graph1, complex_graph2} of AD abnormal circuits were calculated for ADNI2 and BrainLat, respectively. Independently identified circuits in both datasets exhibited highly consistent trends across these three metrics as thresholds varied. Collectively, these results fully demonstrate the application potential of personalized brain functional networks in the study of abnormal neural circuits.

\section*{Discussion}\label{sec3}

\begin{figure*}[htbp]
    \centering
    \includegraphics[width=\textwidth]{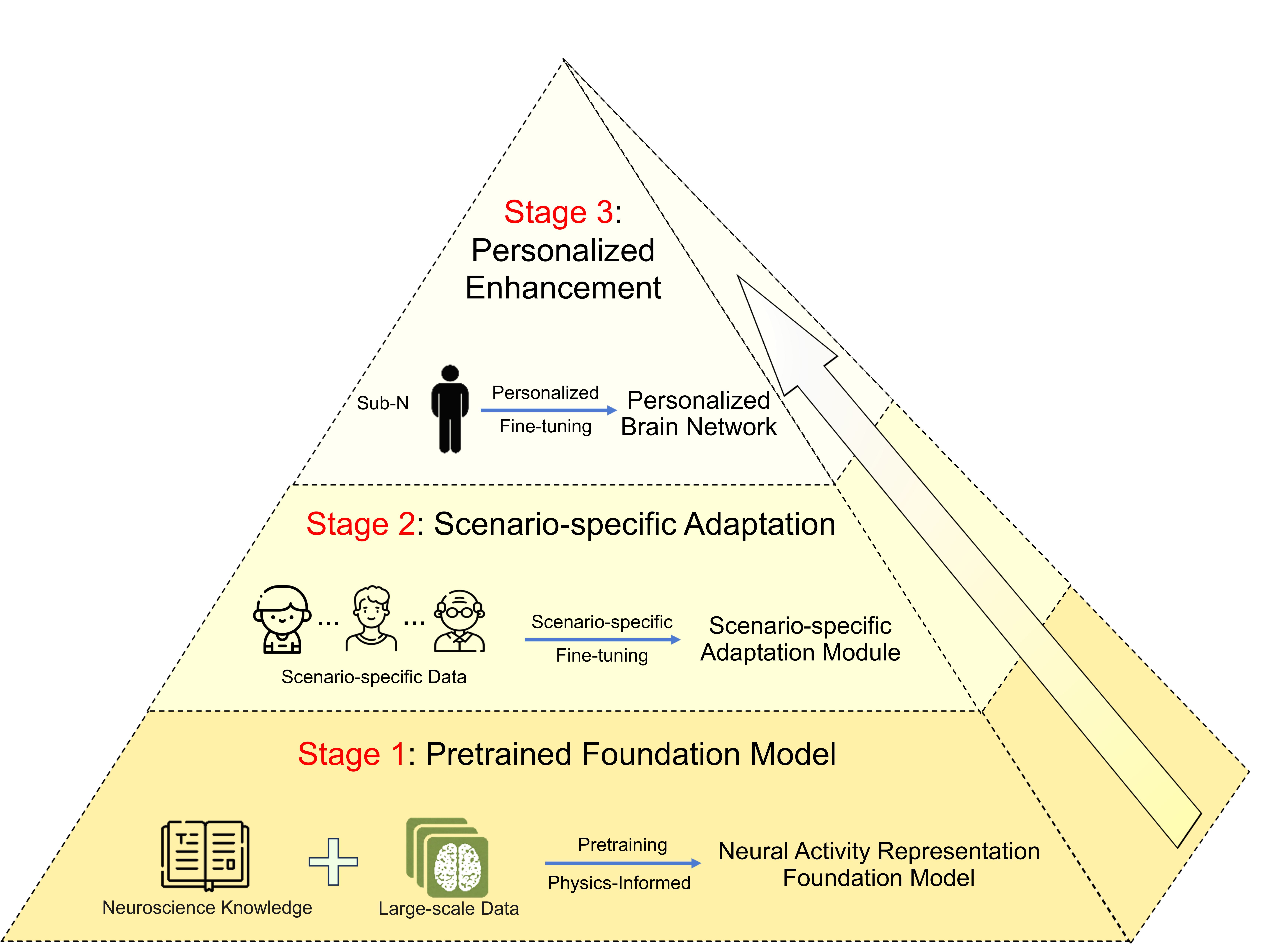}
    \caption{\textbf{Schematic of the "Foundation Model-Personalized Adaptation paradigm" for personalized brain functional networks construction.} The framework expands through three hierarchical stages: Stage 1, Integration of large-scale heterogeneous neuroimaging data and multi-perspective neuroscience knowledge to train a physics-informed neural activity representation foundation model, establishing robust representation capabilities for diverse scenarios. Stage 2, Implementation of scenario-specific adaptation via lightweight adapters, allowing the framework to accurately capture specific scenario features while retaining general representation capabilities. Stage 3, Personalized enhancement utilizing unique subject data to further decouple subject-specific neural activity patterns, finally constructing the personalized brain functional network.}
    \label{fig:6}
\end{figure*}

From a macroscopic perspective, brain activity is a neural dynamic process shaped by brain geometry \cite{anatomic_constrain}. In variable and heterogeneous scenarios across dimensions such as subjects’ disorder types, subjects’ age groups, subjects’ races, and image acquisition strategies, underlying neural dynamics drive significant variations in both the neural activity spatial distribution pattern and the neural activity correlation pattern. However, dominant brain functional network construction methods based on “pre-defined brain atlases” and “linear assumptions” inherently overlook the intrinsic variations of the above neural activity patterns. Thus, dominant methods fail to precisely capture the variations in spatial distribution pattern and correlation pattern of neural activity across heterogeneous scenarios. This is specifically reflected in two aspects: First, dominant methods rely on pre-defined brain atlases derived from ex vivo specimens and group-averaged analysis. But neuroscientific evidence indicates that functional units are task-\cite{functional_unit_is_task_state_subject1}, and subject-dependent\cite{functional_unit_is_task_state_subject1, functional_unit_is_task_state_subject2}. For example, the construction process of the Desikan-Killiany atlas is based on the analysis of ex vivo specimens and the averaging of the analysis results across the group\cite{dk_atlas}. These atlases fail to capture functional boundary variations caused by differences in subjects’ age groups, subjects’ brain disorder types, subjects’ race (e.g., linguistic differences), and image acquisition strategies\cite{Different_Disease_Different_Atlas, age_groups, languages, scan_duration1, scan_duration2}. This makes dominant methods unable to effectively capture variations in neural activity spatial distribution patterns. Second, neural activity correlation estimation relies on linear assumptions (e.g., pearson correlation, partial correlation)\cite{Micapipe, CATO}, which compromise the accurate characterization of neural activity correlations in heterogeneous scenarios. This makes dominant methods unable to effectively capture variations in neural activity correlation patterns. Here, a personalized brain functional network construction framework is proposed. The proposed framework aims to overcome the limitations by representing personalized neural activity patterns across heterogeneous scenarios. The proposed framework follows a pretraining-finetuning paradigm: a foundation model is pre-trained on large-scale fMRI data to extract unified neural activity pattern representations, and subsequently fine-tuned under neural dynamics constraints to obtain personalized representations in heterogeneous scenarios. On one hand, personalized representations of neural activity patterns in specific scenarios were obtained through fine-tuning. On the other hand, the fine-tuning process under neural dynamics constraints embeds dynamic information into personalized representations. This enables a more precise characterization of neural activity patterns. On this basis, the personalized representation-guided parcellation and correlation estimation modules effectively capture dynamic changes in functional boundaries and neural activity correlations, thereby overcoming the limitations of current methods.

Compared to current brain functional network construction methods, the proposed framework exhibits significant advantages in neural modulation, abnormal neural circuit analysis, brain disorder diagnosis, and motor imagery decoding. First, the core bottleneck in current neural modulation lies in the lack of an effective objective basis for target localization, leading to high uncertainty in effectiveness. In the experiment above, brain functional networks constructed by current methods fail to reflect neural activity patterns in heterogeneous scenarios, limiting their effectiveness in clinical neural modulation (the median recovery rate was only 11\% on the NEUROCON dataset). In contrast, the proposed framework obtained a median recovery rate of 26\% on the NEUROCON dataset. This indicates the application potential of personalized brain functional networks in clinical neural modulation. Second, consistency of brain functional networks is a core criterion for evaluating the reliability of a construction framework. However, current methods often suffer from the insufficient-consistency challenge across heterogeneous scenarios. Specifically, the current challenge in identifying abnormal neural circuits lies in the difficulty of maintaining result consistency in variable and heterogeneous scenarios. This results in insufficient reliability of these neural circuits in the analysis of pathological mechanisms. In the experiment above, the median cosine similarity of abnormal neural circuits across three ADHD datasets was 0.64, compared to 0.57 for the baseline. Another challenge is ensuring consistency under short-term, intervention-free conditions. In the experiment above, personalized brain functional networks exhibited significantly higher median one minus PDiv values (0.65–0.8) than those constructed by the baseline method (0.55–0.65). This not only indicates the high reliability of the personalized brain functional network construction framework in heterogeneous scenarios, but also suggests that abnormal neural circuits identified via personalized brain functional networks provide a robust basis for pathophysiological mechanism studies. Third, for downstream tasks such as brain disorder diagnosis, physiological index prediction, and individual identification, the core objective of brain functional network construction is to improve the prediction precision of these tasks. Across these tasks, the proposed method obtained significantly superior performance compared to the baseline method. This indicates that personalized brain functional networks can be widely applied to various diagnostic and predictive tasks.

Limitations of this study primarily lie in the diversity of pre-training data and neural dynamic models. First, the pre-trained foundation model was trained exclusively on the UK Biobank (UKB) dataset\cite{UKB}. Although the UKB dataset is large-scale and includes various brain disorder types, its data distribution exhibits bias. Subjects are predominantly native English speakers aged 40-69, and the fMRI acquisition strategy is limited to a single protocol. This bias may lead to inadequate representation capabilities of the foundation model when encountering data from children, non-native English speakers, or data acquired via different acquisition strategies. Second, the proposed framework for personalized brain functional network construction was solely informed by a wave model. Neural dynamic models from different perspectives were not considered. Models describing neural dynamics include the wave model, the neural mass model, and the the haemodynamic model et al. For instance, the wave model can efficiently capture macroscopic signal propagation patterns determined by brain geometry; the neural mass model describes the interactions between excitatory and inhibitory neuron populations within each brain region through mean-field approximation. Consequently, a single model inevitably embeds dynamic information from only a single perspective into personalized representations. Therefore, integrating neural dynamic models from different perspectives remains a critical challenge in the framework for constructing personalized brain functional networks.

In future work, we plan to establish a "Foundation Model-Personalized Adaptation" paradigm for constructing personalized brain functional networks (Fig. \ref{fig:6}). This paradigm will drive the framework's expansion through three stages. First, we will integrate large-scale neuroimaging datasets covering variable and heterogeneous scenarios (e.g., all age groups, multiple acquisition strategies, diverse races, disorders, and cognitive tasks) to train the neural representation foundation model. This will significantly enhance the model’s representation capacity for neural activity patterns in heterogeneous scenarios. We also plan to adaptively incorporate multi-perspective neuroscience knowledge into the framework, providing domain priors which is difficult to obtain via purely data-driven approaches. Second, we will introduce scenario-specific data and lightweight adapters for scenario adaptation. This allows the framework to retain general representation capabilities while acutely capturing features of specific scenarios. Third, building on scenario adaptation, we will perform personalized representation enhancement to finally construct personalized brain functional networks. Here, the model will utilize unique subject data to further decouple subject-specific neural activity patterns. Such a "Pre-trained Foundation Model-Scenario-Specific Adaptation-Personalized Enhancement" framework is expected to pave the way for future precision diagnosis and treatment of brain disorders.

\section*{Methods}

\subsection*{Datasets} 
The framework for constructing personalized brain functional networks was evaluated on 18 distinct datasets comprising a total of 1,008 subjects. All image acquisitions were approved by the relevant ethics committees and are publicly available for basic research. The included data encompass diverse age groups, image acquisition strategies, and brain disorder types.
Specifically, the data sources are detailed as follows: healthy child and young adult samples were selected from the ABIDE dataset; healthy elderly samples were collected from the ADNI dataset; long-duration scan data (>50 min) were sourced from the HCP dataset; short-duration scan data (<15 min) originated from the GSP dataset; AD-related samples were obtained from the ADNI, and BrainLat datasets; ADHD samples were selected from ADHD200, ADHD-RISA, and UCLA-CNP; ASD samples were sourced from ABIDE I, ABIDE II, and CRF; MDD samples were primarily aggregated from RFDP, BAML, SRAA-MDD, and RfsDP; PD samples were derived mainly from Neurocon, Taowu, and PPMI; data for BMI prediction were obtained from ISTMDCR-S1; and data for motor imagery decoding tasks were selected from the HKMI-fMRI dataset.

\subsection*{Data Preprocessing} 
In this study, fMRI data preprocessing was conducted using DeepPrep \cite{deepprep}, an efficient processing framework constructed upon deep learning algorithms and the Nextflow workflow manager. The pipeline integrates core procedures, including whole-brain anatomical segmentation, cortical surface reconstruction, cortical surface registration, and volumetric spatial normalization; executes standard functional preprocessing operations such as head motion correction, distortion correction, and nuisance regression; and finally resamples the data into the standard space of the fs\_LR 32k surface format.

\subsection*{Personalized brain functional network construction framework} 
The framework for constructing personalized brain functional networks operates through a sequential process: first, personalized representations of neural activity patterns are extracted via a neural dynamics-driven personalized neural activity pattern representation module; subsequently, these personalized representations are employed to sequentially drive the adaptive brain parcellation module and the neural activity correlation estimation module; and ultimately, the personalized brain functional network is constructed.

\subsubsection*{Neural dynamics-driven personalized neural activity pattern representation module}
\textbf{(i) Personalized neural activity pattern representation module.} The framework for constructing personalized brain functional networks extracts personalized neural activity pattern representations from fMRI data via the personalized neural activity pattern representation module. This module is built upon a pre-trained foundation model featuring a latent diffusion model as its core architecture \cite{lt_diffusion}. The architecture comprises three components: an fMRI encoder $\mathcal{E}$, which maps fMRI data into a low-dimensional latent space; an fMRI decoder $\mathcal{D}$, which reconstructs latent features back to the original space; and a diffusion model operating within the latent space. The extraction process proceeds as follows: first, fMRI data are mapped to the latent space via the encoder $\mathcal{E}$; subsequently, during the process where the data are processed by the diffusion model and reconstructed to the original space by the decoder $\mathcal{D}$, intermediate layer features of the model are extracted as the personalized neural activity pattern representation, denoted as ${{\mathbf{Z}}_{pattern}}$. Building upon this, an additional decoder $\mathcal{D}_1$ was trained to map the personalized representation back to the cortical space. This decoder $\mathcal{D}_1$ plays a critical role in multiple subsequent steps, including joint fine-tuning and personalized representation-guided adaptive brain parcellation.

\textbf{(ii) Neural dynamics-driven mechanism.} This framework incorporates a neural dynamics mechanism to fine-tune the aforementioned pre-trained foundation model under the constraints of the neural wave equation \cite{anatomic_constrain}, thereby integrating intrinsic brain dynamic information into the personalized representations. First, by combining the geometric eigenmodes $\{\psi_{1},\dots,\psi_{k},\dots,\psi_{N}\}$ within the fs\_LR 32k space, any neural activity can be represented as
\begin{align}
\phi(\mathbf{r},t)=\sum_{k=1}^{N}a_k(t)\psi_k(\mathbf{r})
\end{align}

By substituting this formulation into the wave model, a constraint equation is derived:
\begin{align}
    % 第一行：微分算子部分
    & \bigg[ \frac{1}{\gamma_s^{2}}\frac{\partial^{2}}{\partial t^{2}} + \frac{2}{\gamma_s}\frac{\partial}{\partial t} + 1 \notag \\ 
    &- r_s^{2}\nabla^{2}\bigg]\sum_{k=1}^{N}a_k(t)\psi_k(\mathbf{r}) = Q(\mathbf{r},t)
\end{align}
where $\nabla^2$ denotes the Laplace operator, describing the spatial diffusion of neural signals across the cortical surface; $\gamma_s$ represents the temporal damping rate, governing the decay speed of neural responses; $r_s$ indicates the spatial correlation length; $Q(\mathbf{r},t)$ signifies the external drive (configured as white noise to simulate resting-state activity); $\mathbf{r}$ denotes the position on the cortical surface; and $t$ represents time.

The problem of capturing neural activity patterns from fMRI data ($\mathbf{B}∈{{\mathbb{ℝ}}^{V\mathrm{×}T}}$, where $V$ denotes the number of voxels and $T$ represents the length of the time series) in specific scenarios is defined as the following optimization problem:
\begin{align}
&\min_{a_k(t)}
\sum_{\mathbf{r}\in\mathrm{fsLR\_32K}}
 \sum_{t=1}^{T}
 \Bigl\|
  \sum_{k=1}^{N}a_k(t)\psi_k(\mathbf{r})-\mathbf{B}_{\mathbf{r},t}
 \Bigr\|^2
\\ \notag
&\text{s.t.}\quad
\biggl[
 \frac{1}{\gamma_s^{2}}\frac{\partial^{2}}{\partial t^{2}}
 +\frac{2}{\gamma_s}\frac{\partial}{\partial t}
 +1 \\ \notag
&-r_s^{2}\nabla^{2}
\biggr]
\sum_{k=1}^{N}a_k(t)\psi_k(\mathbf{r})
=Q(\mathbf{r},t)
\end{align}

By solving the above optimization problem, the neural activity pattern ${{\mathrm{Φ}}_{dyn}}(\mathbf{r,}t)$ is obtained. Based on this pattern, a neural dynamics consistency fine-tuning loss ${\mathcal{L}_{phy}}$ is constructed. This loss is then minimized to enable the foundation model to integrate neural dynamics information into the personalized representations during the extraction process.
\begin{mycustomequation}
\mathcal{L}_{\mathrm{phy}}
=
\mathbb{E}_{\mathbf{r},\,t}
\Bigl\|
\mathcal{D}_{1}(\mathbf{Z}_{\mathrm{pattern}})
-
\Phi_{\mathrm{dyn}}(\mathbf{r},t)
\Bigr\|_{F}^{2}
\end{mycustomequation}

\textbf{(iii) Pre-training strategy.} A foundation model was pre-trained utilizing fMRI data from the UK Biobank \cite{UKB}. This model is designed to extract personalized neural activity pattern representations. Large-scale pre-training enpowers the model with enhanced generalization capabilities across various heterogeneous scenarios. The pre-training process is divided into three stages: the first stage involves the pre-training of the fMRI encoder $\mathcal{E}$ and the fMRI decoder $\mathcal{D}$; the second stage focuses on the pre-training of the diffusion model within the neural activity latent space; and the third stage aims to pre-train the decoder ${\mathcal{D}_1}$ to map the personalized representations back to the cortical space.

In the first stage, the fMRI encoder $\mathcal{E}$ and fMRI decoder $\mathcal{D}$ are pre-trained based on the input fMRI data ($\mathbf{B}∈{{\mathbb{ℝ}}^{V\mathrm{×}T}}$, where $V$ denotes the number of voxels and $T$ represents the length of the time series). To effectively extract intrinsic patterns of brain activity, a signal reconstruction pre-training strategy is adopted. Specifically, this process constructs an autoencoder architecture: the encoder $\mathcal{E}$ is responsible for mapping high-dimensional, redundant voxel-level fMRI signals into a low-dimensional, compact latent space; while the decoder $\mathcal{D}$ utilizes these latent variables to attempt an accurate reconstruction of the original fMRI input. The objective of this pre-training stage is to minimize the mean squared error between the original fMRI signals and the reconstructed signals, with the optimization objective function expressed as:
\begin{align}
\mathcal{L}_{\mathrm{Stage1}}
=
\mathbb{E}_{\mathbf{B}\sim\mathcal{D}_{\mathrm{data}}}
\Bigl[
\bigl\|
\mathbf{B}-\mathcal{D}\bigl(\mathcal{E}(\mathbf{B})\bigr)
\bigr\|_{2}^{2}
\Bigr]
\end{align}

In the second stage, the encoder $\mathcal{E}$, trained in the first stage, is frozen, and the low-dimensional latent features $\mathbf{Z}=\mathcal{E}(\mathbf{B})$ extracted by it are utilized as input to pre-train the diffusion model operating within the latent space. This process comprises two steps: forward diffusion and reverse denoising. The forward process gradually adds Gaussian noise to the authentic neural latent features until these features degrade into pure random noise. The core of the pre-training lies in the reverse process, which involves training a denoising network $\varepsilon_\theta$ to predict and eliminate the noise added at each time step. In this manner, the model learns the probability distribution of neural signals in the latent space. The objective of this pre-training stage is to minimize the difference between the predicted noise and the actually added noise, with the optimization objective function expressed as:
\begin{align}
\scalebox{0.95}{$
\mathcal{L}_{\mathrm{Stage2}}
=
\mathbb{E}_{\mathbf{Z}\sim\mathcal{E}(\mathbf{B}),\,
          \varepsilon\sim\mathcal{N}(\mathbf{0},\mathbf{I}),\,
          t}
\Bigl[
\bigl\|
\varepsilon-\varepsilon_{\theta}(\mathbf{Z}_{t},t)
\bigr\|_{2}^{2}
\Bigr]
$}
\end{align}
where ${\mathbf{Z}_t}$ denotes the latent feature after noise has been added at time step $t$, and $ϵ$ represents the actually sampled standard Gaussian noise.

In the third stage, specifically, the encoder $\mathcal{E}$, decoder $\mathcal{D}$, and diffusion model trained in the previous two stages are frozen, and utilized to extract the personalized neural interaction pattern representation ${{\mathbf{Z}}_{pattern}}$. The core task of this stage is to train the decoder ${\mathcal{D}_1}$ to enable the reconstruction of fMRI signals in the cortical space based on these personalized representations. The objective of this pre-training stage is to minimize the difference between the cortical signals reconstructed by decoder ${\mathcal{D}_1}$ and the real fMRI data, with the optimization objective function expressed as:
\begin{align}
\mathcal{L}_{\mathrm{Stage3}}
=
\mathbb{E}_{\mathbf{B}\sim\mathcal{D}_{\mathrm{data}}}
\Bigl[
\bigl\|
\mathbf{B}-\mathcal{D}_{1}(\mathbf{Z}_{\mathrm{pattern}})
\bigr\|_{2}^{2}
\Bigr]
\end{align}

\subsubsection*{Personalized representation-guided adaptive brain parcellation module}

The brain parcellation task is formulated as a Maximum A Posteriori (MAP) estimation problem based on Markov Random Fields (MRF) over the cortical surface mesh $G=(V, \mathcal{E})$. Here, $V$ denotes the set of mesh vertices, and $\mathcal{E}$ represents the set of mesh edges. This model aims to identify the optimal label configuration $L = \{l_v \mid v \in V\}$, where $l_v \in \{1,\dots,K\}$ signifies the brain region class to which vertex $v$ belongs. The optimal label configuration is obtained by minimizing the following Gibbs energy function \cite{schaefer2018local}:
\begin{align}
&E(\mathbf{L})
= \sum_{v\in V} E_{\mathrm{unary}}(l_v) \\ \notag
&+ \lambda \sum_{(u,v)\in\mathcal{E}} E_{\mathrm{pairwise}}(l_u,l_v)
\end{align}
where $\lambda$ serves as a hyperparameter controlling the smoothing intensity. In the Gibbs energy function, the unary potential ${E_{unary}}$ is modeled as a dual von Mises-Fisher (vMF) distribution to simultaneously capture functional consistency and spatial aggregation. Furthermore, to preserve functional boundaries while smoothing noise, a gradient-weighted Potts model is employed to define the pairwise interaction term ${E_{pairwise}}$.

The minimization of the energy function $E$ is achieved through an iterative optimization strategy guided by the personalized neural activity pattern representation. Specifically, in the initialization phase, the decoder ${\mathcal{D}_1}$ is first utilized to map the personalized neural activity pattern representation onto the cortical space. Subsequently, self-supervised learning based on feature similarity is performed in this space, and the resulting clustering outcome is adopted as the initial label configuration ${\mathbf{L^‘}}$ for the brain parcellation task. This strategy is designed to ensure that the optimization process begins from a reasonable state adaptable to neural activity patterns under variable and heterogeneous scenarios. Following this, the energy function $E$ is minimized by alternating between parameter re-estimation and Graph Cuts-based label optimization to obtain the optimal label configuration ${\mathbf{L}}$.

\subsubsection*{Personalized representation-guided neural activity correlation estimation module}
This module utilizes the personalized neural activity pattern representation to guide the neural activity correlation encoder (Transformer architecture \cite{transformer}) for capturing varying neural activity correlations in heterogeneous scenarios.
First, based on the results of the adaptive brain parcellation, the fMRI voxel time series in each brain region are averaged to obtain a region-level neural activity matrix $\mathbf{H}\in\mathbb{R}^{K\times T}$, where $K$ denotes the number of functional regions of interest (ROIs).
Second, a self-attention mechanism is employed to capture the correlations among the node activity sequences $\mathbf{H}=\{\boldsymbol{h}_1,\boldsymbol{h}_2,\dots,\boldsymbol{h}_K\}$.
In this process, $\mathbf{H}$ is mapped into queries $Q_s$, keys $K_s$, and values $V_s$, calculated as follows:
\begin{align}
\mathbf{H}_{\mathrm{intra}}
= \mathrm{Softmax}\!\left(\frac{Q_s K_s^T}{\sqrt{d_k}} V_s\right)
\end{align}
where $\mathbf{H}_{\mathrm{intra}}=\{\boldsymbol{h}_1^{\mathrm{intra}},\dots,\boldsymbol{h}_K^{\mathrm{intra}}\}$ denotes a set of features, where each element $\boldsymbol{h}_i^{\mathrm{intra}}$ aggregates information from other brain regions and models the correlations of the brain activities.

To adapt to heterogeneous scenarios, a cross-attention mechanism is utilized for feature fusion: the set ${{\mathrm{H}}_{intra}}$ is mapped into query vectors $Q_c$, while the personalized representation ${{\mathbf{Z}}_{pattern}}$ is mapped to key vectors $K_c$ and value vectors $V_c$. Based on this configuration, the model is capable of dynamically retrieving and integrating relevant prior patterns from the personalized representation according to the intrinsic correlation states of the current brain regions.
\begin{align}
\mathbf{H}_{\mathrm{cross}}
= \mathrm{Softmax}\!\left(\frac{Q_c K_c^T}{\sqrt{d_k}}\right)V_c
\end{align}
Through this design, a deep fusion of neural activity correlation estimation and personalized priors is realized. Ultimately, the personalized brain functional network is constructed by calculating the correlation coefficients among the elements of $\mathbf{H}_{\mathrm{cross}}=\{\boldsymbol{h}_1^{\mathrm{cross}},\dots,\boldsymbol{h}_K^{\mathrm{cross}}\}$.

\subsection*{Multi-module joint fine-tuning strategy}

To realize effective adaptation between the personalized neural activity pattern representation module and the brain functional network construction task, three key modules are jointly fine-tuned: the personalized neural activity pattern representation module, the personalized representation-guided adaptive brain parcellation module, and the personalized representation-guided neural activity correlation estimation module.

First, to realize adaptation between the personalized neural activity pattern representation module and the adaptive brain parcellation module, the personalized representation $\mathbf{Z}_{\mathrm{pattern}}$ is mapped via decoder $\mathcal{D}_1$ to the cortical space to obtain $\hat{X}=\{\hat{x}_v\mid v=1,\dots,V\}$. Combining this with the optimal labels $\mathbf{L}$ from the adaptive parcellation, the energy function is transformed into a loss function for joint fine-tuning:
\begin{align}
&\mathcal{L}_{\mathrm{seg}}
= \sum_{v} E_{\mathrm{unary}}(l_v,\hat{x}_v)
+ \\ \notag 
&\lambda \sum_{(u,v)} E_{\mathrm{pairwise}}(l_u,l_v)
\end{align}
Second, the personalized brain functional network obtained from the neural activity correlation estimation module is fed into downstream classification and regression tasks. The task loss is constructed by calculating the difference between the predicted results  and the ground truth labels:
\begin{align}
\mathcal{L}_{\mathrm{task}}
= \mathcal{L}_{\mathrm{CE/MSE}}(\hat{Y},\,Y_{\mathrm{gt}})
\end{align}
The joint fine-tuning loss is defined as follows:
\begin{align}
\mathcal{L}_{\mathrm{total}}
= \mathcal{L}_{\mathrm{task}}
+ \lambda_{1}\mathcal{L}_{\mathrm{seg}}
\end{align}
\subsection*{An open tool for personalized brain functional networks construction} 
An open tool for constructing personalized brain functional networks has been developed. This tool is designed to establish a flexible ecosystem: users may either act as contributors by contributing personalized brain parcellations and neural activity correlation estimation methods to an open community, or perform customizations locally to preserve data privacy. This flexible ecosystem empowers users to rapidly access personalized brain parcellations and neural activity correlation estimation methods specialized for specific scenarios, thus facilitating the efficient and convenient construction of personalized brain functional networks. The tool and detailed instructions are available at:
\url{https://github.com/Leetcode115/A_Personalized_Brain_Functional_Network_Construction_Tool}

\subsection*{Virtual neural modulation}

SHAP analysis is first applied to identify optimal modulation targets. Subsequently, virtual perturbations are applied to these targets to simulate the intervention effects of clinical neural modulation.

\subsubsection*{SHAP-based modulation target identification}

SHAP is a unified framework for interpreting machine learning models \cite{SHAP}. It estimates the contribution of each feature by averaging all marginal contributions to the prediction task. In this study, the SHAP framework is utilized to analyze the diagnostic model for specific diseases. A set of brain regions highly correlated with pathology was identified as modulation targets, denoted as ${S_{target}}$, to ensure the precision of the modulation strategy. The modulation targets are obtained using the following formula: 
\begin{align}
S_{\mathrm{target}}
= \underset{S\subseteq\mathcal{V},\,|S|=K}{\arg\max}
\sum_{i\in S} \phi(i)
\end{align}

\subsubsection*{Neural dynamics learning}

To train a model capable of simulating neural activity, the brain activity fitting problem is formulated as follows \cite{virtual_pertabation}:
\begin{align}
x_t = F_{\theta}(x_{t-3},\,x_{t-2},\,x_{t-1}) + \varepsilon_t
\end{align}
where $x_t$ represents the neural activity state of brain regions at time $t$, and $\varepsilon_t$ denotes the Gaussian noise with zero mean and unit variance. The loss function employed to optimize the model's parameters in the simulation of neural activities is formulated as the Mean Squared Error between the predicted and target neural activities:
\begin{align}
&L = \frac{1}{n}\sum_{i=1}^{n}
\bigl(F_{\theta}(x_{t-3},x_{t-2},x_{t-1})\\ \notag 
&+\varepsilon_t-\hat{x}_t\bigr)^{2}
\end{align}
where ${{\hat{x}}_t}\mathrm{\text{ }}$ denotes the ground truth and $n$ represents batch size.
A neural activity model is obtained by minimizing the loss $L$.

\subsubsection*{Virtual perturbation}

To perform virtual neural modulation, virtual perturbations are applied to a set of optimal targets using the neural activity model to obtain the modulated neural activity states. Let $V \in {1, 2, \dots, N}$ denote the set of whole-brain nodes, where $N$ represents the number of nodes and $x^i$ represents the neural activity state of node $i$. The virtual perturbation for node $i$ is implemented using the following formula \cite{virtual_pertabation}:
\begin{align}
&x_t^{\mathrm{pert}}
= F_{\theta}\bigl(x_{t-3}^{i},\,
              x_{t-2}^{i},\,
              x_{t-1}^{i}+\Delta\!\cdot\!e_i\bigr)
+ \varepsilon_t \\ \notag
&, i\in S_{\mathrm{target}}
\end{align}
where $\Delta$ denotes the perturbation intensity, and $e_i$ represents a Gaussian noise vector with a length equal to $x_{t−1}^i$.

\subsubsection*{Consistency evaluation of brain functional networks}

To quantify the consistency of personalized brain functional networks, Portrait Divergence defined subsequently is adopted as the evaluation metric. This metric encodes the distribution information of all shortest paths in the network by constructing a network portrait, thereby evaluating full-scale features ranging from local connectivity to global connectivity \cite{PDiv}.

For a given brain network $G$, the network portrait matrix $B$ is constructed. The element ${B_{l,k}}$ is defined as the number of nodes in the network that have $k$ nodes at a shortest-path distance $l$:
\begin{align}
    &B_{l,k} = \left| \Bigl\{ v \in V \Bigm| \right. \\ \notag   % 用 \right. 假装结束
    &\left. \mathrm{count}\bigl(\{u\in V\mid d(v,u)=l\}\bigr)=k \Bigr\} \right| % 用 \left. 假装开始
\end{align}
where $d(v,u)$ denotes the shortest-path distance between nodes $v$ and $u$, $0\leq l \leq {d_{max}}$(network diameter), and $0 \leq k \leq N-1$($N$ is the total number of nodes in the network).

To compare portraits of different brain networks, the matrix $B$  is normalized into a probability distribution $P(k,l)$. This distribution represents the joint probability of randomly selecting two nodes at distance $l$, where one node has $k$ neighbors at that distance. The formula is as follows:
\begin{align}
P(k,l)
= \frac{1}{N}\,B_{l,k}\;
  \frac{1}{\sum_{c}n_{c}^{2}}
  \sum_{k'=0}^{N} k' B_{l,k'}
\end{align}
where $n_c$ denotes the number of nodes in the $c$-th connected component of the network, and the denominator term is used to normalize the total number of weighted paths.

Based on the above probability distribution, the Jensen-Shannon divergence is used to measure the topological discrepancy between two networks. The final Portrait Divergence is defined as:
\begin{align}
&\mathrm{PDiv}(G_1,G_2)
= \frac{1}{2}\mathrm{KL}(P\parallel M) + \\ \notag
&\frac{1}{2}\mathrm{KL}(Q\parallel M)
\end{align}
where $P$ and $Q$ are the portrait probability distributions of networks $G_1$ and $G_2$, respectively; $M=(P+Q)/2$ is the mixture distribution; and $KL(⋅||⋅)$ denotes the Kullback-Leibler divergence.
Under this definition, the result ranges between [0,1], with 1 indicating perfect overlap of the portrait distributions and 0 indicating completely distinct topological structures.

\section*{Data availability}

The ABIDE dataset is available at \url{https://fcon_1000.projects.nitrc.org/indi/abide/}. The ADNI dataset is available at \url{https://adni.loni.usc.edu/data-samples/adni-data/}. The BrainLat dataset is availiable at \url{https://github.com/PunksB1602/BrainLat_dataset}. The HCP dataset is available at \url{https://db.humanconnectome.org}. The GSP dataset is available at \url{https://www.neuroinfo.org/gsp}. The ADHD200 dataset is available at \url{https://fcon_1000.projects.nitrc.org/indi/adhd200/}. The ADHD-RISA dataset is available at \url{https://openneuro.org/datasets/ds003500}. The UCLA-CNP dataset is available at \url{https://openneuro.org/datasets/ds000030}. The CRF dataset is available at \url{https://openneuro.org/datasets/ds002522}. The RFDP dataset is available at \url{https://openneuro.org/datasets/ds002748}. The BAML dataset is available at \url{https://openneuro.org/datasets/ds000171}. The SRAA-MDD dataset is available at \url{https://openneuro.org/datasets/ds003770}. The RfrsDP dataset is available at \url{https://openneuro.org/datasets/ds004101}. The Neurocon and Taowu datasets are available at \url{https://fcp-indi.s3.amazonaws.com/data/Projects/INDI/umf_pd}. The PPMI dataset is available at \url{https://www.ppmi-info.org/}. The ISTMDCR-S1 dataset is available at \url{https://openneuro.org/datasets/ds000229}. The HKMI-fMRI dataset is available at \url{https://openneuro.org/datasets/ds003612}. The NBVSC-SCH and NBVSC-BD dataset are available at \url{https://openneuro.org/datasets/ds005073}. Source data are provided with this paper.

\section*{Code availability}
The proposed brain functional network construction framework is available at \url{}.

\backmatter

%\bibliographystyle{sn-nature.bst}
% \bibliography{sn-bibliography}

% \clearpage
%\unsetvruler % 关闭行号
\bibliography{sn-bibliography}% common bib file

\section*{Acknowledgements}\label{sec8}

\section*{Author contributions}\label{sec9}

\section*{Competing interests}\label{sec10}
The authors declare no competing interests.

\setcounter{table}{0}
\renewcommand{\tablename}{}
\renewcommand{\thetable}{Extended Data Table \arabic{table}}

\setcounter{figure}{0}
\renewcommand{\figurename}{Extended Data Fig.}  % 重新定义图片名称
\renewcommand{\thefigure}{\arabic{figure}}  % 保持编号的顺序

\section*{Supplement}

\end{document}